\documentclass[conference]{IEEEtran}
\IEEEoverridecommandlockouts
\usepackage{cite}
\usepackage{amsmath,amsthm,amssymb,amsfonts}
\usepackage{algorithmic}
\usepackage{graphicx}
\usepackage{textcomp}
\usepackage{xcolor}
\usepackage{booktabs} 
\usepackage[linesnumbered,ruled,vlined]{algorithm2e}
\usepackage{svg}
\usepackage{amsmath}
\usepackage{float}
\usepackage[font=small,labelfont=bf]{caption}
\usepackage[utf8]{inputenc}

\usepackage{multirow}
\usepackage{multicol}
\usepackage{makecell}
\usepackage[font=footnotesize]{subfig}
\usepackage{rotating}
\usepackage{enumitem}
\usepackage{tabularx}
\usepackage[normalem]{ulem}

\theoremstyle{plain}
\newtheorem{thm}{Theorem}
\theoremstyle{definition}
\newtheorem{defn}[thm]{Definition} 

\SetAlFnt{\small}

\def\BibTeX{{\rm B\kern-.05em{\sc i\kern-.025em b}\kern-.08em
    T\kern-.1667em\lower.7ex\hbox{E}\kern-.125emX}}
\begin{document}

\title{\huge Opportunistic Air Quality Monitoring and Forecasting with Expandable Graph Neural Networks}

\author{\IEEEauthorblockN{Jingwei Zuo, Wenbin Li, Michele Baldo and Hakim Hacid}
\IEEEauthorblockA{Technology Innovation Institute, Abu Dhabi, UAE \\
Email: $\{$firstname.lastname$\}$@tii.ae}
}

\maketitle

\begin{abstract}
Air Quality Monitoring and Forecasting has been a popular research topic in recent years. 
Recently, data-driven approaches for air quality forecasting have garnered significant attention, owing to the availability of well-established data collection facilities in urban areas. 
Fixed infrastructures, typically deployed by national institutes or tech giants, often fall short in meeting the requirements of diverse personalized scenarios, e.g., forecasting in areas without any existing infrastructure. Consequently, smaller institutes or companies with limited budgets are compelled to seek tailored solutions by introducing more flexible infrastructures for data collection. 
In this paper, we propose an expandable graph attention network (EGAT) model, which digests data collected from existing and newly-added infrastructures, with different spatial structures.
Additionally, our proposal can be embedded into any air quality forecasting models, to apply to the scenarios with evolving spatial structures. The proposal is validated over real air quality data from PurpleAir.
\end{abstract}


\section{Introduction}
Air quality forecasting using data-driven models has gained significant attention in recent years, thanks to the proliferation of data collection infrastructures such as sensor stations and advancements of telecommunication technologies. These infrastructures are typically managed by national institutes (e.g., AirParif\footnote{https://www.airparif.asso.fr/}, EPA\footnote{https://www.epa.gov/air-quality}) or large companies (e.g., PurpleAir\footnote{https://www2.purpleair.com/}) that specialize in air quality monitoring or forecasting services and products. Leveraging existing data collection infrastructures proves beneficial for initial research exploration or validating product prototypes.
However, reliance on fixed infrastructures presents practical constraints when customization is required for specific tasks. For instance, certain monitoring areas may be inadequately covered or completely absent from the existing infrastructures, or the density of coverage may not be sufficient. This issue particularly affects small or mid-sized industrial and academic players who face budget limitations that prevent them from investing in their own infrastructure from scratch, but have specific customization needs.

In addition to data collection, air quality forecasting models trained solely with data from public fixed infrastructures may not perform well for users' specific scenarios, such as forecasting at a higher spatial resolution. Deploying additional sensors as a cost-effective solution can enrich the data and improve forecasting performance without the need to build infrastructures from scratch. 
Subsequently, this targeted solution leads us to consider the practical question: \textit{how we can make use of the data collected from existing infrastructures, when integrating new sensor infrastructures?} 

\begin{figure}
\centering
\includegraphics[width=1\linewidth]{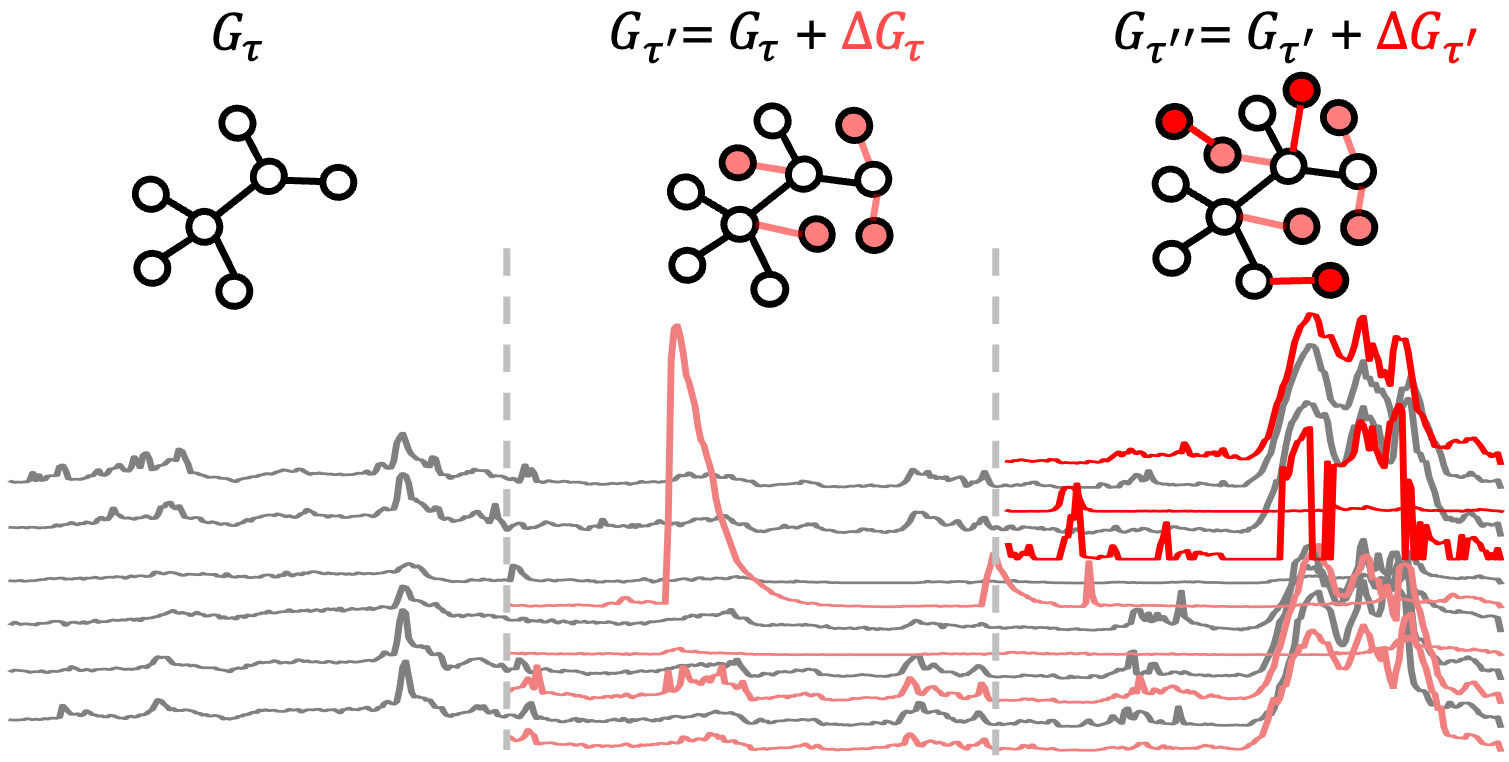}
\caption{Expanded sensor networks and the related $PM_{2.5}$ data at different time. The data was collected with $Purpleair API$~\cite{purpleair}.}
\label{fig:research_background}
\vspace{-1em}
\end{figure}

As depicted in Figure \ref{fig:research_background}, the topological sensor network may change as the urban infrastructure evolves, resulting in varying network structures of air quality sensors. The data collected from the network $G_{\tau}$ needs to be augmented with enriched data from newly installed sensors $\Delta G_{\tau'}$ and $\Delta G_{\tau''}$. Training a model solely on recent data with $G_{\tau''}$ would overlook valuable information contained in the historical data with $G_{\tau}$ and $G_{\tau'}$.

In this paper, we propose an expandable graph attention network (EGAT) that effectively integrates data with various graph structures. This approach is versatile and can be seamlessly embedded into any existing air quality forecasting model. Furthermore, it applies to scenarios where sensors are not installed, enabling accurate forecasting in such areas.
We summarize our approach's main advantages as follows:
\begin{itemize}

    \item \textbf{Less is more:} With fewer installed sensors, we can directly predict the air quality of other unknown area where sensors are not installed and achieve comparable performance to models relying on extensive data collection infrastructures with more sensors.
    \item \textbf{Continual learning with self-adaptation:} The proposed model enables continuous learning from newly collected data with expanded sensor networks, demonstrating self-adaptability to different topological sensor networks.
    \item \textbf{Embeddable module with scalability:} The proposed module can be seamlessly integrated into any air quality forecasting model, enhancing its ability to forecast in real-world scenarios.

\end{itemize}

The rest of this paper starts with a review of the most related work. Then, we formulate the problems of the paper. Later, we present in detail our proposal, which is followed by the experiments on real-life datasets and the conclusion.

\section{Related Work}
\subsection{Air Quality Forecasting}
Data-driven models for air quality forecasting has gained a huge popularity recently. Recent work \cite{zuo2023graph,liang2022airformer} studies graph-based representations of the air quality data by considering the sensor network as a graph structure, which extracts decent structural features between sensor data from a topological view. The air quality forecasting can be then formulated as a spatio-temporal forecasting problem.

Works like DCRNN~\cite{li2018diffusion}, STGCN~\cite{yu2018spatio} and Graph WaveNet~\cite{wu2019graph}, have shown promising results in traffic forecasting tasks. These models can be adapted to air quality forecasting tasks owing to the shared spatio-temporal features present in the data. 
However, in practice, the above-mentioned models often overlook the evolving nature of sensor networks as more data collection infrastructures are incrementally built. Consequently, these models require re-training from scratch on the most recent data that reflects the evolved sensor network. It may result in the loss of valuable information contained in outdated data collected from different network configurations.


\subsection{Expandable Graph Neural Networks}

In the field of graph learning, several works, such as ContinualGNN \cite{wang2020streaming} and ER-GNN \cite{zhou2021overcoming}, have incorporated the concept of Continual Learning to capture the evolving patterns within graph nodes.
While these approaches are valuable, it is important to consider spatio-temporal features in air quality forecasting tasks.
Designed for traffic forecasting, TrafficStream~\cite{chen2021trafficstream} considers evolving patterns on both temporal and spatial axes; ST-GFSL~\cite{lu2022spatio} introduces a meta-learning model for cross-city spatio-temporal knowledge transfer. However, these works primarily focus on shared (meta-)knowledge between nodes, and give less attention to expandable graph structures. 
Basically, spectral-based graph neural networks (GNNs) face challenges when scaling to graphs with different structures due to the complexity of reconstructing the Laplacian matrix. To address this issue, our paper explores the use of spatial-based GNNs, such as Graph Attention Networks (GAT) \cite{velivckovic2018graph}, for expandable graph learning in air quality forecasting tasks.




\section{Problem Formulation}
\begin{defn}(Air Quality Forecasting). Given an air quality sensor network $G=\mathcal{\{V, E\}}$, where $\mathcal{V}=\{v_1, ..., v_N\}$ is a set of $N$ sensor nodes/stations and $\mathcal{E}=\{e_1, ..., e_E\}$ is a set of $E$ edges connecting the nodes, the air quality data $\{AQI_{t}\}_{t=1}^{T}$ and meteorological data $\{M_{t}\}_{t=1}^{T}$ are collected over the $N$ stations, where $T$ is current timestamp. We aim to build a model $f$ to predict the $AQI$ over the next $T_p$ timestamps.
\end{defn}

To simplify, we denote input data as $\mathcal{X}$= $\{AQI_{t}, M_{t}\}_{t=1}^{T}$
= $\{x_t\}_{t=1}^{T}\in \mathbb{R}^{N \times F \times T }$. 
Each node contains $F$ features representing $PM_{2.5}$, $PM_{10}$, humidity, temperature, etc. As $PM_{2.5}$ is \textit{most reported and most difficult-to-predict}~\cite{yi2018deep}, we take $PM_{2.5}$ as the AQI prediction target $\mathcal{Y}$=$\{y_t\}_{t = T+1}^{T+T_{p}} \in \mathbb{R}^{N\times T_{p}}$. 

\begin{defn}(Expanded Sensor Network). Given a sensor network at $\tau$: $G_{\tau}$ = $\{\mathcal{V}_{\tau}, \mathcal{E}_{\tau}\}$ with $N_{\tau}$ sensors, the network at $\tau'$: $G_{\tau'}$=$G_{\tau}$+$\Delta G_{\tau}$ = $\{V_{\tau'}, E_{\tau'}\}$ expands $G_{\tau}$ to $N_{\tau'}$ sensors. 
\end{defn}

We aim to build a model $f$, which is firstly trained over a dataset $\{\mathcal{X}_{\tau}\}$ on a sensor network $G_{\tau}$ = $\{\mathcal{V}_{\tau}, \mathcal{E}_{\tau}\}$, and can be incrementally trained over $\{\mathcal{X}_{\tau'}\}$ on an expanded network $G_{\tau'}$. 
For inference, given a sequence $\mathcal{X} \in \mathbb{R}^{N_{\tau'} \times F \times T}$ and a sensor network $G_{\tau'}$, the model $f$ can predict the $AQI$ for the next $T_{p}$ time steps $\mathcal{Y}$=$\{y_t\}_{t = T+1}^{T+T_{p}} \in \mathbb{R}^{N\times T_{p}}$, where $N_{\tau'} \geq N_{\tau}$. 
\section{Our proposals}

\begin{figure*}[!htbp]
\centering
\includegraphics[width=1\linewidth]{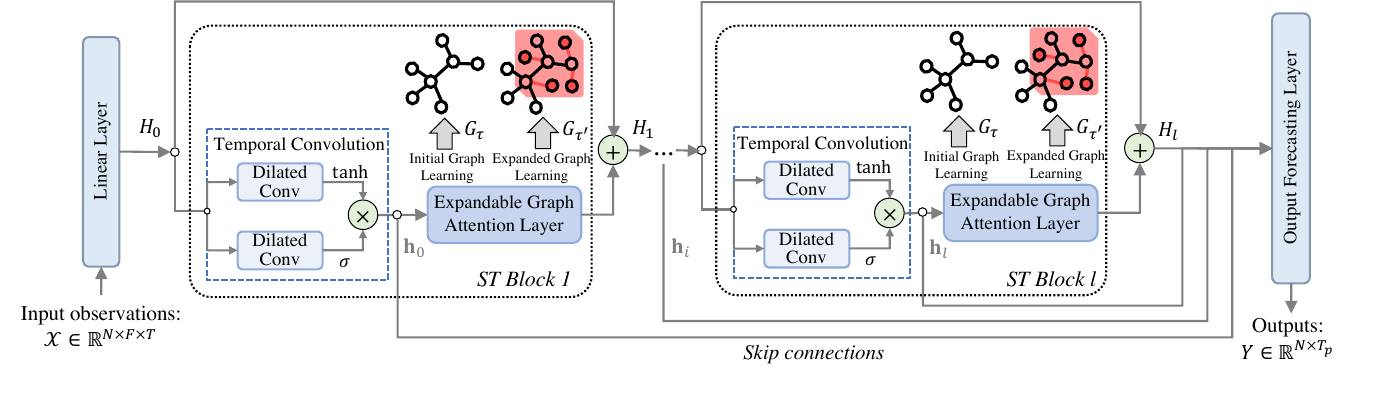}
\vspace{-2em}
\caption{Global system architecture of EGAT}
\label{fig:system_structure}
\vspace{-1em}
\end{figure*}

In this paper, we adopt Graph WaveNet~\cite{wu2019graph} as the backbone model, which consists of $l$ Spatio-Temporal (ST) Blocks. However, our proposed EGAT can be integrated to any spatio-temporal models with adaptations on graph network layers. We employ Temporal Convolution Network (TCN) to encode the temporal dynamics of the AQIs. Specifically, as shown in Figure~\ref{fig:system_structure}, we designed an Expandable Graph Attention Network (EGAT) to learn from the data with evolving graph structures. 
The output forecasting layer takes skip connections on the output of the final ST Block and the hidden states after each TCN module for final predictions.


\subsection{Temporal Dynamics with Temporal Convolution Network}
Compared to RNN-based approaches, Temporal Convolution Network (TCN)~\cite{wu2019graph} allows handling long-range sequences in a parallel manner, which is critical in industrial scenarios considering the model efficiency. 

Given an input air quality sequence embedding $H$= $f_{linear}(\mathcal{X}) \in \mathbb{R}^{N \times d \times T}$, a filter $\mathcal{F}\in \mathbb{R}^{1 \times \mathrm{K}}$, $\mathrm{K}$ is the temporal filter size, $\mathrm{K}=2$ by default. The dilated causal convolution operation of $H$ with $\mathcal{F}$ at time $t$ is represented as:
\begin{equation}
\small
    H \star \mathcal{F}(t)= \textstyle\sum_{s=0}^{\mathrm{K}}\mathcal{F}(s) H(t-\textbf{d} \times s) \in \mathbb{R}^{N \times d\times T'}
\end{equation}
where $\star$ is the convolution operator, $\textbf{d}$ is the dilation factor, $d$ is the embedding size, $T'$ is the generated sequence length.
We define the output of a gated TCN layer as:
\begin{equation}
\small
    \textbf{h} = tanh(W_{\mathcal{F}^{1}} \star H) \odot \sigma(W_{\mathcal{F}^{2}} \star H) \in \mathbb{R}^{N \times d \times T'}
\end{equation}
where $W_{\mathcal{F}^{1}}$, $W_{\mathcal{F}^{2}}$ are learnable parameters, $\odot$ is the element-wise multiplication operator, $\sigma(\cdot)$ denotes Sigmoid function.

\subsection{Expandable Graph Attention Networks (EGATs)}
Graph attention network (GAT)~\cite{velivckovic2018graph}, as a weighted message-passing process, models neighboring nodes' relationships via their inherent feature similarities.
Given a set of air pollution features at time $t$: $\textbf{h}(t)$ = $\{h_1, h_2, ..., h_N\}, h_i\in \mathbb{R}^{N \times d}$ as input of a graph attention layer, following~\cite{velivckovic2018graph}, we define the attention score between node $i$, $j$ as: 
\begin{equation}\label{eq:attention_score}
\small
\alpha_{i j}=\frac{\exp 
\left(
\operatorname{a}\left(W h_i, W h_j\right)
\right)
}{\sum_{k \in \mathcal{N}_i} \exp \left(
\operatorname{a}\left(W h_i, W h_k\right)
\right)} 
\end{equation}

where $W \in \mathbb{R}^{d \times d'}$ is a weight matrix, $\operatorname{a}$ is the attentional mechanism as mentioned in~\cite{velivckovic2018graph}: $\mathbb{R}^{d'} \times \mathbb{R}^{d'} \to \mathbb{R} $, and $\mathcal{N}_i$ is a set of neighbor nodes of $v_i$.
A \textit{multi-head attention} with a nonlinearity $\sigma$ is employed to obtain abundant spatial representation of $v_i$ with features from its neighbor nodes $\mathcal{N}_i$:
\begin{equation}
\label{eq:multi_head_GAT}
\small
h_{i}'=\sigma\left(\frac{1}{K} \sum_{k=1}^K \sum_{j \in \mathcal{N}_i} \alpha_{i j} W^k h_j\right)
\end{equation}
Therefore, the GAT layer in $i$-th ST Block can be defined as:
\begin{equation}\label{eq:gat}
\small
H_{i+1} = \sigma \left(\frac{1}{K} \sum_{k=1}^K \mathcal{A} \textbf{h}_{i} W^k \right)
\end{equation}
where $\mathcal{A}$=$\{\alpha_{ij}\} \in \mathbb{R}^{N \times N}$, $H_{i+1} \in \mathbb{R}^{N \times d' \times T}$, $W^k \in \mathbb{R}^{d \times d'}$.

When expanding the graph with new sensor nodes, we scale up the GAT layers on new nodes while conserving the information learned over the old ones. Basically, new nodes can be considered during both model's training and inference.


\subsubsection{Expandable Graph Network Training}
We consider that the sensor network expands with the newly built infrastructures. The model learned from $G_{\tau}$ can be updated with recent data over $G_{\tau'}$ without re-training the model from scratch. 

From Equation \ref{eq:gat}, with new embeddings $\mathbf{h}_{\tau'}$$\in$$\mathbb{R}^{N_{\tau'} \times d \times T}$, the weight matrix $W^{k}$ stays unchanged; only the adjacency matrix requires updates: $\mathcal{A}_{\tau} \in \mathbb{R}^{N_{\tau} \times N_{\tau}}$ $\to$ $\mathcal{A}_{\tau'}\in \mathbb{R}^{N_{\tau'} \times N_{\tau'}}$.
We re-define $\mathcal{N}_i$=$\{\mathcal{N}_{i,\tau}, \mathcal{N}_{i,\tau'}\}$ as the $k$ nearest neighbors of $v_{i}$, where $\mathcal{N}_{i,\tau}$ denotes neighbors from existing nodes, $\mathcal{N}_{i,\tau'}$ indicates those from newly added nodes. 
Given a set of new sensors $\Delta \mathcal{V}_{\tau}$, we obtain new edge connections $\Delta \mathcal{E}_{\tau}$=$\{\mathcal{N}_{i}\}_{i=1}^{\Delta N}$, where $\Delta N$=$N_{\tau'}-N_{\tau}$, with $\mathcal{O}(N_{\tau'} \Delta N)$ time for distance computations.
According to Equation~\ref{eq:attention_score}, the attentional mechanism will apply to $\Delta \mathcal{E}_{\tau}$ with $\mathcal{O}(\Delta N k)$ time. Therefore, the attention score between node $i, j$ can be re-defined as:
\begin{equation}\label{eq:attention_score_new}
\footnotesize
\alpha_{i j}=\frac{\exp 
\left(
\operatorname{a}\left(W h_i, W h_j\right)
\right)
}{\sum\limits_{k \in \mathcal{N}_{i,\tau}} \! \exp \left(
\operatorname{a}\left(W h_i, W h_k\right)
\right)
\!+\!
\sum\limits_{k \in \mathcal{N}_{i,\tau'}} \! \exp \left(
\operatorname{a}\left(W h_i, W h_k\right)
\right)
} 
\end{equation}
In this manner, we can update the graph layer, i.e., $\mathcal{A}_{\tau'}$ incrementally by considering cached attention scores over $\mathcal{E}_{\tau}$, reducing the time complexity to $\mathcal{O}(N_{\tau'} \Delta N + \Delta N k)$. 
This is much faster than rebuilding the entire graph layer ($\mathcal{O}(N_{\tau'}^{2})$).
\subsubsection{Expandable Graph Network Inference}
When no sensors are installed in (unseen) areas, \textit{Spatial Smoothing} can be performed on the unseen node $v_i$. Based on its spatial location, we incorporate predictions from its neighbor nodes:
\begin{equation}
    Y_i = \sum_{j\in \mathcal{N}_i} a_{ij} Y_j, N_i = \{v_j | dist(v_i, v_j) < \varepsilon \}
\end{equation}
where $\mathcal{N}_i$ is the first-order neighbors of $v_i$ (excluding $v_i$, as the data on $v_i$ is unavailable), $a_{ij}=1-\frac{dist(v_{i}, v_{j})}{\sum_{k\in \mathcal{N}_{i}} dist(v_{i}, v_{k})}$ is the inverse Euclidean Distance (ED) between $v_i$ and $v_j$, $\varepsilon$ is a threshold which decides the neighboring sensor nodes. 

We propose a robust \textit{Spatial Representation Smoothing} technique that considers richer spatial relationships, in the embedding space, between unseen and existing nodes. Given an unseen node $v_i$, its embedding $h_i$ can be defined as follows:
\begin{equation}
\small
h_{i}=\sigma\left(\frac{1}{K} \sum_{k=1}^K \sum_{j \in \mathcal{N}_i} a_{i j} W^k h_j\right )
\end{equation}
where $a_{ij}$ is the inverse ED between $v_i$ and $v_j$, $W^{k}$ is the learned weights in each attention head as shown in Equation~\ref{eq:multi_head_GAT}.

\subsection{Output Forecasting Layer}
For final predictions, we take skip connections as shown in~\cite{wu2019graph} on the final ST Block's output and hidden states after each TCN.
The concatenated output features are defined as:
\begin{equation}
\small
    O = (\textbf{h}_{0} W^{0} + b^{0})\| ... \| (\textbf{h}_{l-1} W^{l-1} + b{l-1}) \ \| (\mathcal{H}_{l} W^{l} + b^{l})
\end{equation}
where $O \in \mathbb{R}^{N\times (l+1)d}$, $W_{s}^{i}$, $b_{s}^{i}$ are learnable parameters for the convolution layers. Two fully-connected layers are added to project the concatenated features into the desired dimension:
\begin{equation}
\small
    \hat{\mathcal{Y}} = (ReLU(OW_{fc}^{1} +  b_{fc}^{1})) W_{fc}^{2}  +  b_{fc}^{2} \in \mathbb{R}^{N\times T_{p}}
\end{equation}
where $W_{fc}^{1}$, $W_{fc}^{2}$, $b_{fc}^{1}$, $b_{fc}^{2}$ are learnable parameters. We use mean absolute error (MAE) \cite{wu2019graph} as loss function for training.

\section{Experiments}
In this section, we demonstrate the effectiveness of EGAT with real-life air quality datasets. The experiments were designed to answer the following questions:

\begin{itemize}
    
    \item[\textbf{Q1}] \textit{Continual learning with self-adaptation:} How well can our model make use of the ancient data with different graph structures, to improve the model's performance? 

    \item[\textbf{Q2}] \textit{Flexible Inference on unknown areas:} How well is our model at predicting air quality in areas without any sensors installed? i.e., no available data over these areas. 

\end{itemize}

\subsection{Experimental Settings}
\subsubsection{Dataset description}
We base our experiments on real air quality data~\cite{zuo2023unleashing} collected via PurpleAir API~\cite{purpleair}, which contains the AQIs and meteorological data in San Francisco (within $10 \, km^2$) between 2021-10-01 and 2023-05-15. The datasets are split to training, validation, test sets with \textit{7:1:2}. 
Table \ref{AQIDatasets} shows more details of the collected datasets. For PurpleAirSF-1H, we adopt the last 12-hour data to predict the AQI (i.e., PM2.5) for the next 12 hours. For PurpleAir-6H, we consider the last 72 hours to predict the next 72 hours.

\begin{table}[!htbp]
\centering
\caption{Summary statistics of PurpleAirSF-1H/6H}
\label{AQIDatasets}
\scalebox{0.85}{
\begin{tabular}{ccccccc}
\toprule
\textbf{Data} & \textbf{\#Nodes} & \textbf{\#Features} & \textbf{Sampling} & \textbf{Observations} & \textbf{Missing} \\
\midrule
PurpleAirSF-1H  & 112              & 19                           & 1 hour               & 29 011 024            & 1.566\%                \\
PurpleAirSF-6H  & 232              & 19                           & 6 hours              & 10 054 648            & 1.231\%  \\
\bottomrule
\end{tabular}
}
\end{table}

\subsubsection{Execution and Parameter Settings}
We take Graph WaveNet as the backbone model. However, our proposal can be integrated to any air quality forecasting models. All the tests are done on a single Tesla A100 GPU of 40 Go memory. The forecasting accuracy of all tested models is evaluated by three metrics~\cite{zuo2023graph}: mean absolute error (MAE), root-mean-square error (RMSE) and mean absolute percentage error (MAPE).

\subsubsection{Baselines}
We compare EGAT with various model variants and with Graph WaveNet~\cite{wu2019graph}:

\begin{itemize}
    \item \textbf{GraphWaveNet} (GWN)~\cite{wu2019graph}: Trained on expanded graph data, as it is non-adaptable to different graph structures.
    \item \textbf{EGAT-Rec}: EGAT trained on data with expanded graph; 
    \item \textbf{EGAT-FI-SS}: EGAT trained on data over ancient graph, Flexible Inference (FI) with \textit{Spatial Smoothing} is applied; 
    \item \textbf{EGAT-FI-SRS}: EGAT trained on ancient data, FI with \textit{Spatial Representation Smoothing} is employed; 
    \item \textbf{EGAT}: EGAT trained on both ancient and recent data.
    
\end{itemize}


\subsection{Experimental Results}
Table \ref{table:results_expand_node_ratio} and Table \ref{table:results_expand_time_ratio} reports the average errors (12/72H) regarding the expanding node ratio and expanding time ratio determined by the deployment.
\textbf{Bold} values indicate the best results, while \uline{underlined} values represent the second-best.

EGAT consistently outperforms other models in continual learning with different node ratios and time radios, owning to its ability to leverage rich data from various graph structures. While GWN performs better than EGAT-Rec, this can be attributed to the k-order diffusion process in GCN. Even so, EGAT surpasses GWN by incorporating ancient graph data, further validating our proposal in graph adaptations (\textbf{Q1}).

When forecasting in unknown areas, EGAT-FI-SS provides approximate AQIs through \textit{Spatial Smoothing}. However, its performance deteriorates with a high number of expanded nodes due to spatial sparsity. EGAT-FI-SRS performs better than EGAT-FI-SS and sometimes even better than GWN and comparable to EGAT, validating the viability of \textit{Spatial Representation Smoothing} for unknown areas' prediction (\textbf{Q2}).

\begin{table}[t]
\centering
\caption{Performance comparison regarding different ratios of \textbf{expanded nodes}, we fix the time ratio with expanded nodes as 10\%.}
\label{table:results_expand_node_ratio}
\scalebox{0.68}{
\begin{tabular}{clp{0.7cm}p{0.7cm}p{0.8cm}p{0.7cm}p{0.7cm}p{0.8cm}p{0.7cm}p{0.7cm}p{0.8cm}}
\toprule
                &       & \multicolumn{3}{c}{Expand node = 10\%}                                   & \multicolumn{3}{c}{Expand node = 20\%}                                    & \multicolumn{3}{c}{Expand node = 40\%}                                  \\
                
          & Models                & MAE                 & RMSE          & MAPE(\%)        & MAE                 & RMSE          & MAPE(\%)       & MAE                 & RMSE          & MAPE(\%)         \\
\midrule
\multirow{3}{*}{\rotatebox{90}{P.AirSF-1H}}    & Graph WaveNet               & \uline{3.62}       & \uline{10.77}  & \uline{10.76}  & \uline{3.60}       & \uline{10.83}  & \uline{10.00}  & \uline{3.62}       & \uline{10.85}  & \uline{10.61}  \\
 & EGAT-Rec                    & 3.83               & 11.02          & 12.51          & 3.76               & 10.96          & 11.02          & 3.86               & 11.09          & 12.18          \\
 & EGAT-FI-SS                  & 4.60               & 14.69          & 19.30          & 5.76               & 16.50          & 19.69          & 8.18               & 20.82          & 48.49          \\
 & EGAT-FI-SRS                 & 3.88               & 11.23          & 12.41          & 4.01               & 12.32          & 13.44          & 4.65               & 13.21          & 16.12          \\
 & \textbf{EGAT}          & \textbf{3.47}      & \textbf{10.73} & \textbf{7.73}  & \textbf{3.56}      & \textbf{10.82} & \textbf{8.47}  & \textbf{3.45}      & \textbf{10.76} & \textbf{8.78}          \\

\midrule
\multirow{3}{*}{\rotatebox{90}{P.AirSF-6H}}    & Graph WaveNet               & 6.65               & \uline{16.73}  & 31.47          & \uline{6.65}       & \uline{16.51}  & \uline{23.93}  & \uline{7.04}       & \uline{19.62}  & \uline{25.65}  \\
 & EGAT-Rec                    & 7.90               & 23.32          & 24.74          & 9.27               & 25.66          & 32.51          & 8.41               & 24.56          & 28.92          \\
 & EGAT-FI-SS                  & 6.34               & 22.40          & 26.24          & 8.42               & 29.20          & 36.74          & 11.10              & 35.26          & 80.79          \\
 & EGAT-FI-SRS                 & \uline{5.85}       & 17.21          & \uline{22.41}  & 7.21               & 22.45          & 26.85          & 9.45               & 26.12          & 31.21          \\
 & \textbf{\textbf{EGAT}} & \textbf{5.46}      & \textbf{13.96} & \textbf{18.69} & \textbf{5.18}      & \textbf{13.83} & \textbf{16.21} & \textbf{5.24}      & \textbf{13.85} & \textbf{19.23}       \\

\bottomrule
\end{tabular}
}
\end{table}

\begin{table}[t]
\centering
\caption{Performance comparison regarding different ratios of \textbf{time with expanded nodes}, we fix the expanded node ratio as 10\%.}
\label{table:results_expand_time_ratio}
\scalebox{0.68}{
\begin{tabular}{clp{0.7cm}p{0.7cm}p{0.8cm}p{0.7cm}p{0.7cm}p{0.8cm}p{0.7cm}p{0.7cm}p{0.8cm}}
\toprule
                &       & \multicolumn{3}{c}{Expand time = 10\%}                                   & \multicolumn{3}{c}{Expand time = 20\%}                                    & \multicolumn{3}{c}{Expand time = 40\%}                                  \\
                
          & Models                & MAE                 & RMSE          & MAPE(\%)        & MAE                 & RMSE          & MAPE(\%)       & MAE                 & RMSE          & MAPE(\%)         \\
\midrule
\multirow{3}{*}{\rotatebox{90}{P.AirSF-1H}}    & Graph WaveNet               & \uline{3.62}       & \uline{10.77}  & \uline{10.76}  & \uline{3.57}       & \uline{10.74}  & \uline{8.91}   & \uline{3.45}       & \textbf{10.75} & \uline{7.03}   \\
 & EGAT-Rec                    & 3.83               & 11.02          & 12.51          & 3.77               & 10.96          & 11.76          & 3.54               & 10.78                  & 8.87           \\
 & EGAT-FI-SS                  & 4.60               & 14.69          & 19.30          & 4.84               & 14.85          & 26.88          & 4.23               & 12.09                  & 17.50          \\
 & EGAT-FI-SRS                 & 3.88               & 11.23          & 12.41          & 4.01               & 12.32          & 14.33          & 3.60               & 11.12                  & 9.56           \\
 & \textbf{EGAT}          & \textbf{3.47}      & \textbf{10.73} & \textbf{7.73}  & \textbf{3.42}      & \textbf{10.60} & \textbf{7.43}  & \textbf{3.41}      & \uline{10.77}          & \textbf{6.61}            \\

\midrule
\multirow{3}{*}{\rotatebox{90}{P.AirSF-6H}}    & Graph WaveNet               & 6.65               & 16.73          & 31.47          & \uline{5.71}       & \uline{14.35}  & \uline{18.20}  & 5.40               & 14.21                  & 17.45          \\
 & EGAT-Rec                    & 7.90               & 23.32          & 24.74          & 6.38               & 16.36          & 19.33          & 6.10               & 14.99                  & 18.11          \\
 & EGAT-FI-SS                  & 6.34               & 22.40          & 26.24          & 7.00               & 25.85          & 23.99          & 6.72               & 21.90                  & 26.70          \\
 & EGAT-FI-SRS                 & \uline{5.85}       & 17.21          & \uline{22.41}  & 6.22               & 16.43          & 19.22          & \uline{5.23}       & \uline{14.11}          & \uline{14.26}  \\
 & \textbf{\textbf{EGAT}} & \textbf{5.46}      & \textbf{13.96} & \textbf{18.69} & \textbf{4.85}      & \textbf{13.82} & \textbf{13.87} & \textbf{4.78}      & \textbf{13.98}         & \textbf{12.18}           \\

\bottomrule
\end{tabular}
}
\vspace{-2em}
\end{table}

\section{Perspectives and Conclusion}
In this paper, we propose an Expandable Graph Attention Network (EGAT) for Air Quality monitoring and forecasting. It incorporates historical and recent graph data, which prevents industrial players with budget limitations from investing in their own infrastructures from scratch. EGAT also allows predicting air quality in areas without installed sensors. Future work includes comparing additional expandable graph learning models and exploring transfer learning and node alignment techniques to reduce re-training effort in industrial scenarios.


\bibliographystyle{IEEEtran}
\bibliography{references.bib}
\end{document}